\documentclass[conference]{IEEEtran}
\IEEEoverridecommandlockouts
\usepackage{cite}
\usepackage{amsmath,amssymb,amsfonts}
\usepackage{algorithmic}
\usepackage{graphicx}
\usepackage{textcomp}
\usepackage{paralist, tabularx}
\usepackage[colorlinks=true, allcolors=blue]{hyperref}
\usepackage{subcaption}
\usepackage[svgnames]{xcolor}

\definecolor{dgreen}{rgb}{0.0,0.6,0.0}

\def\BibTeX{{\rm B\kern-.05em{\sc i\kern-.025em b}\kern-.08em
    T\kern-.1667em\lower.7ex\hbox{E}\kern-.125emX}}
\begin{document}

\title{Why is the prediction wrong? Towards underfitting case explanation via meta-classification
\thanks{}
}

\author{%
  \IEEEauthorblockN{%
    Sheng ZHOU\IEEEauthorrefmark{1}\IEEEauthorrefmark{2},
    Pierre BLANCHART\IEEEauthorrefmark{1},
    Michel CRUCIANU\IEEEauthorrefmark{2} and
    Marin FERECATU\IEEEauthorrefmark{2}%
  }%
  \IEEEauthorblockA{\IEEEauthorrefmark{1} 
  Université Paris-Saclay, CEA, List, F-91120, Palaiseau, France
  }%
  \IEEEauthorblockA{\IEEEauthorrefmark{2} Conservatoire National des Arts et Métiers (CNAM), CEDRIC, 75003, Paris, France
  }%
  Email: sheng.zhou@cea.fr, pierre.blanchart@cea.fr, michel.cru\-cianu@cnam.fr, marin.ferecatu@cnam.fr
}

\maketitle

\vspace{2ex}

\begin{abstract}

In this paper we present a heuristic method to provide individual explanations for those elements in a dataset (data points) which are wrongly predicted by a given classifier. Since the general case is too difficult, in the present work we focus on faulty data from an underfitted model. First, we project the faulty data into a hand-crafted, and thus human readable, intermediate representation (meta-representation, profile vectors), with the aim of separating the two main causes of miss-classification: the classifier is not strong enough, or the data point belongs to an area of the input space where classes are not separable.
Second, in the space of these profile vectors, we present a method to fit a meta-classifier (decision tree) and express its output as a set of interpretable (human readable) explanation rules, which leads to several target diagnosis labels: data point is either correctly classified, or faulty due to a too weak model, or faulty due to mixed (overlapped) classes in the input space.
Experimental results on several real datasets show more than 80\% diagnosis label accuracy and confirm that the proposed intermediate representation allows to achieve a high degree of invariance with respect to the classifier used in the input space and to the dataset being classified, i.e.\ we can learn the meta-classifier on a dataset with a given classifier and successfully predict diagnosis labels for a different dataset or classifier (or both).

\end{abstract}

\section{Introduction}
\label{sec:introduction}

Recent machine learning models for prediction have attained excellent performance in many data classification tasks, and are widely applied in various scenarios where prediction is needed (e.g.\ medical diagnostics or financial analysis). However, as applications begin to concern 
every aspect of our daily life, there is an increased need for transparency and interpretability of the model decision process, in order to earn user's trust and reveal the reasoning behind model decision in terms comprehensible by humans \cite{zhang21interp}.

Indeed, interpretability of machine learning algorithms has grown to become an important topic in recent years, especially due to the black box nature of many ML algorithms \cite{guidotti2018survey}, aspect which is even more important when they fail: users need to understand the causes of failures such as to avoid them in the future. The concept of interpretability for machine learning aims at presenting model predictions in a human-understandable way, or providing information to interpret the predictions or the failures \cite{doshi2017towards}.

Modern ML models (such as XGBoost \cite{chen2016xgboost} or artificial neural networks \cite{hastie2009elements}) are powerful enough to fit the most complex patterns in data distributions with rather contained computation costs but, unfortunately for most of them, the decision-making process is not transparent to users (black-box behavior). In such a case it is important to have tools to investigate their behavior, especially when they fail, such as to be able to detect general patterns of model's behavior on its faulty predicting data, if they exist \cite{arrieta2020xai}.

In this direction, we propose in this work an analysis and diagnostic pipeline to interpret the behavior of an ML model in the input space, around data points where it fails to give the expected prediction.  
More precisely, the task is to analyse and \textit{diagnose faulty individual points}, where the label predicted by the model differs from the ground truth. More specifically, because it is difficult to approach the problem with full generality, we restrict ourselves to the more amenable situation when the model is \textit{underfitted}. Recall that underfitting is a global property of a model which summarizes the fact that the accuracy, as a classification performance measure, fails to achieve high values on both training and test sets; furthermore, there is no big gap in accuracy between the train and test sets.   In this situation, given a faulty data point (wrongly classified), our aim is to reach one of the two following diagnostics :

\begin{itemize}
\item \textbf{The model is too weak}: the model fits poorly the true data distribution, because its \textit{capacity as well as complexity is not high enough} so it fails to get sufficient information on the true decision boundary (model underfitting). An example of this case is trying to fit a linear classifier on a dataset that is separable, e.g., by a parabola. In this case, the solution is to use a more complex (and possibly more expensive) classifier, able to fit the data.

\item \textbf{Data classes are mixed-up}: the data point belongs, in the input space, to an area where several class labels mix up beyond \textit{statistical resolution}, meaning that different training samples would give very different separating boundaries. In this case it is impossible to generalize from one sample to the other (e.g.\ from training to test), even using a very strong classifier, implying that the data in this area \textit{cannot be separated properly} in terms of their labels. To do better in this case, the user needs to devise more powerful features for the data, to resolve the ambiguity.
\end{itemize}

To achieve this, we project original feature space $X$ into a hand-crafted intermediate representation space $Z$ ($x\rightarrow z$), where the representation $z$ of a point $x$ encodes the local behavior of the classifier $C(x)$ in the $X$ space around the point $x$, compared to what its output should be (the ground-truth). We hand-craft the variables defining the $Z$ space, instead of learning them from data, because we want them to be meaningful to a human user, thus easy to interpret.
In the second stage we fit a decision tree in the $Z$ space, to obtain a meta-classifier $M(z(x))$ capable of assigning a diagnostic label to any data point $x\in X$: "data point is well classified" (no classification error), "model is too weak", or "point is in a mixed-up data region".

To summarize, the main contributions of our proposal are as follows :
\begin{itemize}
    \item We propose an intermediate representation of the input data which is model agnostic and data agnostic, that is it does not use specific information about the type of classifier or data being employed.
    \item The proposed approach is end-to-end interpretable: it uses a human readable representation (hand-crafted features) and an interpretable meta-model (decision tree) to help the user understand the behavior of a black-box classifier around the data points where it fails to function properly.
\end{itemize}

\begin{figure*}[!htbp]
    \centering
    \includegraphics[width=0.90\textwidth]{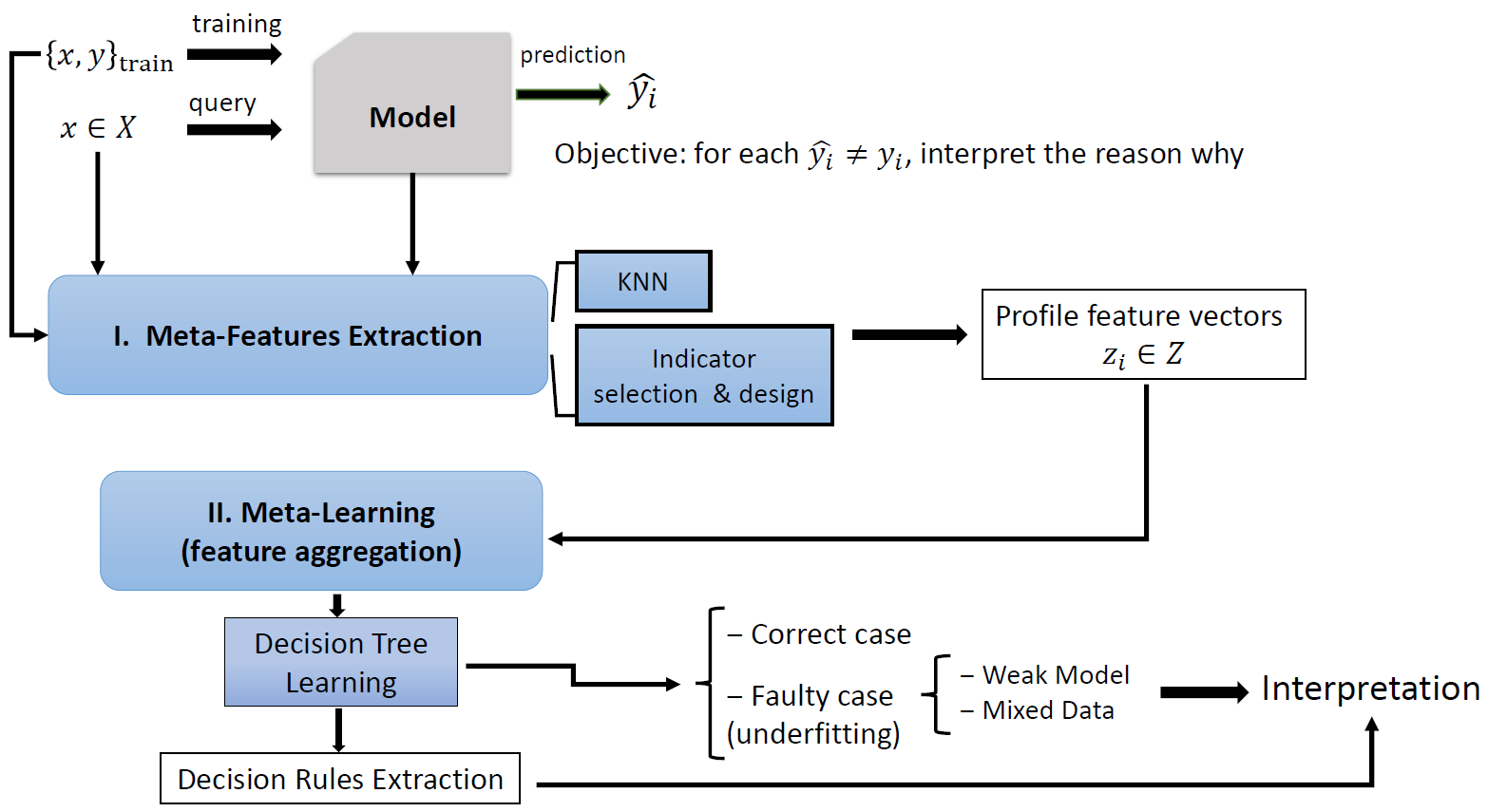}
    \caption{ Main pipeline of our proposal: a black-box model is trained with the train set $\{x_i,y_i\}_\textrm{train}$, the goal being to predict whether the sample is miss-classified and achieve, for each miss-classified sample $x$ (for which $y \neq \hat{y}$), an understanding of the cause of miss-classification: "the model is too weak" or "data classes are mixed-up".
    To achieve this, each data point $x$ is projected by the Meta-Features Extraction module into a profile vector $z$ consisting of several statistical indicators; then the collection of all acquired profile vectors is used by the Feature Aggregation module to train a decision tree on the three classes mentioned above. This will help the user obtain a diagnosis result, understand the reason behind the faulty case, and thus arrive at a practical solution for post-hoc treatment.}
\label{fig:pipeline}
\end{figure*}

 By employing our method the user will know if she needs to put more effort into improving the classifier or into the data harvesting activity, and also have a clear view on why this is the case.
 We test our framework on several large real datasets. The results show systematically more than 80\% accuracy for the meta-classifier (in the $Z$ space), confirming that the proposed intermediate representation permits to achieve a reasonably high degree of invariance with respect to the classifier used in the input space and also to the dataset being classified. That is, we can learn the meta-classifier on an initial collection of classification problems (couples dataset/classifier) and successfully predict diagnosis labels for a different couple dataset/classifier. The workflow of our method is shown in Fig.~\ref{fig:pipeline}.

The paper is organized as follows: in Sec~\ref{sec:relatedwork} we present related work and position our proposal with respect to these, in Sec.~\ref{sec:proposal} we present our framework followed by the experimental validation on several real world datasets in Sec.~\ref{sec:experiments} and we conclude in Sec.~\ref{sec:conclusion} with a synthesis of our achievements and a discussion of future work.

\section{Related Work}
\label{sec:relatedwork}
In this section we present a brief synthesis of the existing state of the art on the topic of diagnosis and interpretability of faulty cases from validation/test data, and position our work with respect to these.

\subsection{Model validation and debugging}

Several existing works focus on machine learning model validation, data analysis and debugging, which serve a similar objective as ours:

\textit{SecureMLdebugger}~\cite{han2020securemldebugger} presents a model debugger by accessing metadata such as model's hyper-parameters, training epochs, evaluation measures and network layouts, without using specific user data to ensure privacy during the analysis. The idea of exploiting model's metadata for debugging and investigation is quite pervasive in the field; we also make use of it but in a different way, i.e.\ to form a set of meta-features characterizing the model, and develop interpretability based on them.

\textit{Slice Finder}~\cite{chung2019slice} is a tool for slicing datasets to obtain certain data subgroups which are identified as problematic to the model. The goal is to provide a more granular view and analysis of model behavior: investigating such kind of problematic data helps explain model poor performance. A slice is defined as a conjunction of feature-value (rule) pairs, in which the number of rule pairs should be few enough to ensure human readability, each acquired slice should have significant impact to affect model performance, and also large enough coverage on validation data. 

\textit{MLCube}~\cite{kahng2016visual} proposes a tool allowing the user to define instance subsets in form of feature-condition conjunctions, and explore the aggregate statistics (like accuracy) or evaluation metrics over these subsets. 
\cite{pastor2021looking} also investigate data subgroups via similar feature-value conjunctions: they identify trouble data subgroups by measuring a divergence metric defined on false positive or false negative rates using Shapley values.

The fore-mentioned methods share a common pathway, that is they assess certain data subgroups via feature-value conjunction structures, which actually match the basic split rules used in decision trees --- they suit thus well the 
logic of tree-based models. Their drawback comes from their nature: the interpretability power of these methods depends on the simplicity of the slice, meaning the granularity of investigation and interpretation is limited by the complexity of the acquired rule set. Unlike them, our method is based on training instances, thus it does not face the issue of rule set complexity but rather draws inferences from data-local population. 

Another drawback when using these methods to solve the model validation and debugging problem is that none of them integrates prior knowledge of model diagnosis categories: their aim is partially the identification of faulty/problematic data, without further analyzing the reason behind the faulty cases. In contrast, our method relies on a clarification of the relationship between the data distribution and the model prediction pairs, and ends up providing interpretations on the reason why certain cases lead the model to make wrong predictions.

\subsection{Data Complexity recognition}
As we already discussed, the mismatch between data distribution in terms of classes and model prediction labels does not always come from model fitting error only, but also from the complexity of the classification problem with respect to the intrinsic distribution of the data.

In this sense, surveys like\cite{ho2002complexity}, \cite{lorena2019complex} list several measures of data distribution complexity in terms of decision boundary dividing different classes. This geometrical point of view leads to three grand classes of measures: linear classifier based (such as minimized error, error rate 
by Linear Programming); nearest neighbor based (fraction of points on boundary by Minimum Spanning Tree method, ratio of average intra/inner class nearest neighbour (NN) distance, error rate and nonlinearity of 1NN classifier); geometry or topology based (maximal Fisher discriminant ratio, overlap region volume, maximal feature efficiency, fraction of points with adherence subsets retained, average number of samples per dimension) \cite{lorena2019complex}. Some of these 
measures focus on data separability or mixture identification, some on individual feature value overlapping, others examine the geometrical properties of the decision boundary such as space covering and non-linearity. While we are not using these results directly, we are taking inspiration from them in devising our meta-features (see next section).

\section{Method Proposal} 
\label{sec:proposal}

\subsection{Problem Formulation} 
\label{sec:formulation}
In this study we only consider binary classification problems. A classifier $C$ is classifying a point $x \in X$ (usually $\mathbb{R}^d$) in one of two classes, $0$ or $1$, i.e.\ $C(x) = y \in \left\{0, 1\right\}$.
Our purpose is to explain why a classifier fails to classify a test point in the right class in the special case when the model is known to be underfitted. We refer to these points as ``faulty'' 
and to the other points as ``normal''. For each faulty point, we further diagnose the cause of classification error 
by considering two main types of behavior:
\begin{itemize}
    \item The model is too simple, i.e.\ there exists a larger capacity 
    model (i.e.\ with more parameters) of the same type that can correctly classify the faulty data considered, while achieving a superior cross-validation performance. 
    \item The faulty data belongs to a region of space where classes $0$ and $1$ overlap, i.e.\ we cannot find a more complex model that correctly classifies the faulty data while achieving superior cross-validation performance. This can happen for instance if the features $x$ are too weak, i.e.\ not providing enough  discriminant information regarding the classification problem at hand. We refer to this case as ``mixed-up'' data in the following. 
\end{itemize}

Our goal is to train a meta-classifier $F$ which maps the test data to three output classes: a normal data is assigned to the class \textbf{\textit{Good Prediction}}, a faulty data is assigned either to the class \textbf{\textit{Weak Model}} or to the class \textbf{\textit{Data Mixed-Up}}.
The meta-classifier takes as input a set of meta-features related to the faulty input point $x$. Meta-features are computed using both $x$, and training points belonging to a neighborhood of $x$ along with their labels.

\subsection{Meta-features extraction} 
\label{sec:metafeatures}

Simple meta-features characterizing local configurations are extracted around a point of interest using $K$-nearest neighbors (KNN). A value for $K$, found by trial and error, of $0.05\times\textrm{(dataset size)}$ provided good results in our case. Training data is then used to compute label-based indicators inside these neighborhoods. In particular, in a neighborhood we extract:
\begin{itemize}
    \item The local model prediction accuracy.
    \item The two-class confusion matrix.
    \item A confidence in the prediction: if the model prediction $\hat{y}_i$ is the probability of 
    class 0, let
    \begin{equation}
    \text{Conf}(x_i) = \frac{\left|\hat{y}_i - 0.5\right|}{0.5} \in [0, 1]\nonumber
    \end{equation}
    then the higher this value, the more confident the model is considered to be in its prediction.
    \item The average confidence in the neighborhood, which reflects the local model confidence around the point of interest.
    \item The average distance to ``allies'', i.e.\ the points from the neighborhood with the same label as $x_i$:
    \begin{equation}
    D_\text{ally}(x_i) = \frac{1}{\vert A_i \vert} \sum_{x_j \in A_i} d\left(x_i, x_j \right) \nonumber
    \end{equation} where $A_i = \left\{x_j \in \text{KNN}\left( x_i \right) | y_j = y_i \right\}$
    \item The average distance to ``opponents'', i.e.\ the points from the neighborhood with a different label than that of 
    $x_i$:
    \begin{equation}
    D_\text{opp}(x_i) = \frac{1}{\vert O_i \vert} \sum_{x_j \in O_i} d\left(x_i, x_j \right) \nonumber
    \end{equation} where $O_i = \left\{x_j \in \text{KNN}\left( x_i \right) | y_j \neq y_i \right\}$.
\end{itemize}

Two other meta-features not using KNN neighborhoods but still characterizing data configurations at a local scale are also extracted:
\begin{itemize}
    \item A Minimum Spanning Tree (MST) based feature described in \cite{ho2002complexity}. This feature requires to compute the MST of the training dataset using a distance criterion to keep only connections between close points. This feature characterizes class separability at a local scale. It is computed as the number of opponents of $x_i$ connected to $x_i$ over the total number of points connected to $x_i$. As such, it is valued between 0 and 1. A value close to zero corresponds to a configuration where the point lies in an homogeneous region of ally points. A value close to $0.5$ corresponds to a configuration where the point lies in a non-homogeneous region made both of ally and opponent points (mixed-up data).
    \item A tree-based meta-feature that can only be added when the binary classifier is a tree ensemble model (other meta-features are model agnostic). This feature is inspired from \cite{tan2020tree} where it is referred to as \textit{Tree space prototype measure}. Given a query point $x_i$ and a point $x_j$ in which to compute the measure, it is obtained as the number of leaves of the tree ensemble model that contain both $x_i$ and $x_j$. This feature is dependent on model complexity, i.e.\ the bigger the number of trees in the ensemble model, the higher it tends to be on average. When averaged over the $K$ nearest neighbors, it quantifies to what extent the $K$ nearest neighbors fall in the same leaves as the query point. As such, it is an indirect measure of model complexity in the considered neighborhood.
\end{itemize}

\subsection{Training the meta-classifier}

The previously described meta-features are aggregated to form profile vectors $z_i \in Z$ associated with each test data. A three-class meta-classifier under the form of a decision tree is trained on these meta-features/profile vectors. The output classes are the ones described in Sec.~\ref{sec:metafeatures}. The choice of a decision tree is justified here by the aim to keep the meta-classifier decisions interpretable. In particular, decision rules can be extracted by following the path between the root and a leaf of the tree. The decision tree selects in a greedy way the most discriminative characteristics for the classification problem at hand and provides an importance score for each of these characteristics. The extracted rules give a more detailed explanation of the meta-classifier decision, all the more since the meta-features are themselves interpretable.

Once the meta-classifier is trained, it can be employed to understand the behavior of other classifiers around the points where they fail to predict the correct label, that is to decide if they need a more powerful classifier or more discriminant features. For this, the user should first extract the meta-features for the dataset on which the investigated classifier was trained.

\section{Experimental evaluation} 
\label{sec:experiments}

In this section, we present experimental validation of our method on several binary classification problems (datasets). To do this we start by pre-processing each dataset in order to generate the ground-truth labels needed for the meta-classifier (Sec.~\ref{sec:resgt}). By employing this ground truth we train the meta-classifier and use prediction accuracy, precision and recall as a measures of success (Sec.~\ref{sec:resmeta}). We also show examples of output interpretations, and we discuss the strengths and weaknesses of our approach.

\subsection{Meta-classifier ground truth generation}
\label{sec:resgt}

Suppose we are given a dataset $D$ associated to a binary classification problem. Each data point is seen as a vector $x$ in the input feature space $X$. Suppose also that we trained a classifier $C$ on some part of $D$ (the training set) and we test on the rest of the dataset (the test set).

Each input point $x\in X$ has a meta-feature vector $z\in Z$ associated with it, computed by the procedure described in Sec.~\ref{sec:metafeatures}.
To be able to train the meta-classifier $F$ in the $Z$ space we need to attach to each element in the input space $x\in X$ one diagnosis label (with respect to the classifier $C$): ``Good Prediction'', ``Weak Model'', ``Data Mixed-Up''. This is obvious for correctly classified data points, but for the miss-classified ones we need to ensure the cause of miss-classification is guaranteed to be the one associated with the label. We describe below the procedure we use to generate these labels.

We start by choosing a strong classifier, for example an XGBoost model with a very large number of trees and large depth. We then iterate several times the following procedure: 

\begin{enumerate}
    \item Randomly split the initial dataset $D$ into two equal disjoint parts, $D_1$ and $D_2$.
    \item Train the classifier with $D_1$ as training set to obtain model $C_1$. Similarly, train the classifier with $D_2$ as training set to obtain model $C_2$.
    \item Remove from $D_1$ the elements miss-classified by $C_2$, obtaining the set ${D'_1}$, and remove from $D_2$ the elements miss-classified by $C_1$,  obtaining the set $D'_2$.
    \item Consider $D = {D'_1} \cup {D'_2}$ and repeat the whole procedure until there are no more miss-classified elements by $C_1$ and $C_2$ (or very few, e.g.\ an accuracy of 99.9\% for both).
\end{enumerate}

In practice, we found that two-three iterations are enough in the procedure above to obtain classifiers with very high accuracy (99.9\%) on the remaining data. The procedure cleans up the dataset, so we are sure that the remaining data can be correctly classified by a model of very high capacity (thus, classes are not mixed-up in the regions covered by this data). 

\begin{figure*}[htbp]
\begin{minipage}[t]{0.49\linewidth}
\centering
\includegraphics[width=8cm]{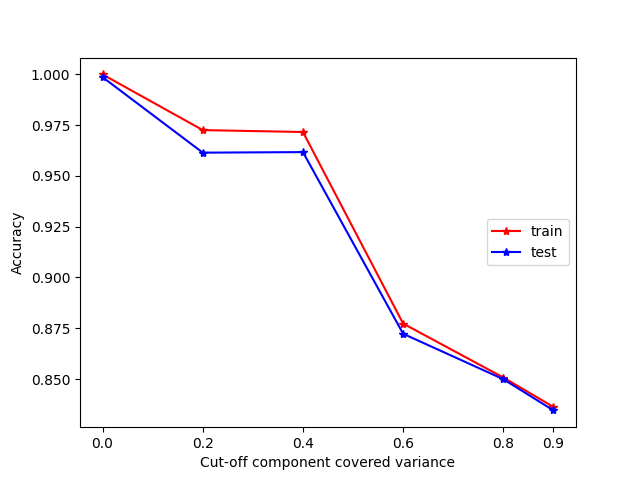}
\subcaption{Spotify, XGB model}
\end{minipage}
\begin{minipage}[t]{0.49\linewidth}
\centering
\includegraphics[width=8cm]{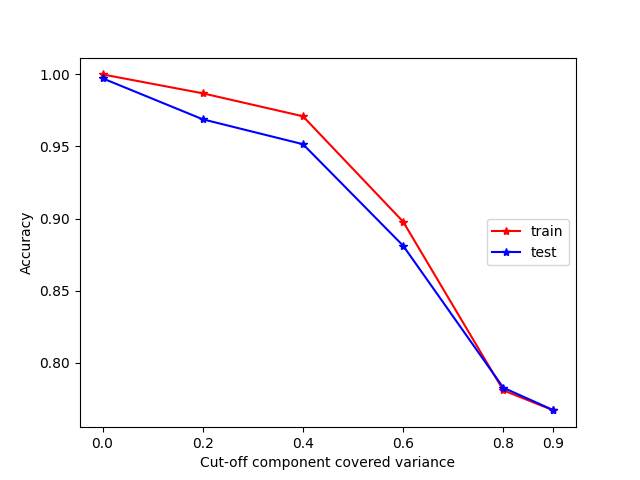}
\subcaption{Spotify, MLP model}
\end{minipage}
\begin{minipage}[t]{0.49\linewidth}
\centering
\includegraphics[width=8cm]{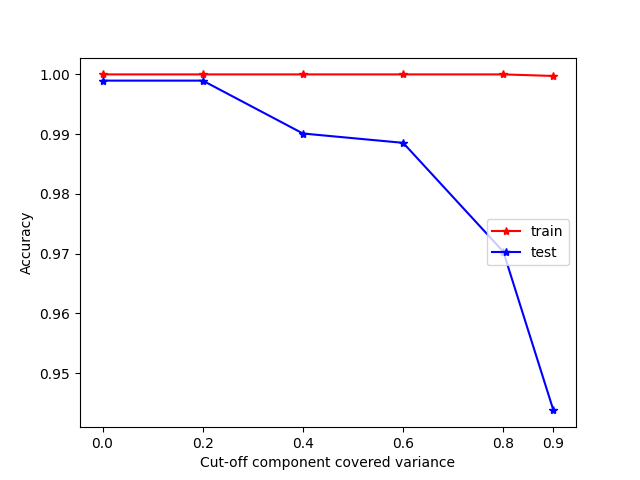}
\subcaption{Fashion MNIST, XGB model (170 trees, depth 6)}
\end{minipage}
\begin{minipage}[t]{0.49\linewidth}
\centering
\includegraphics[width=8cm]{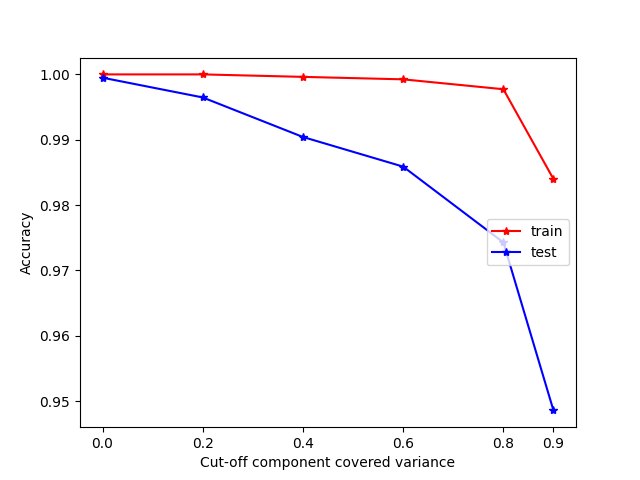}
\subcaption{Fashion MNIST, MLP model (hidden layer size = 50)}       
\end{minipage}
\begin{minipage}[t]{0.49\linewidth}
\centering
\includegraphics[width=8cm]{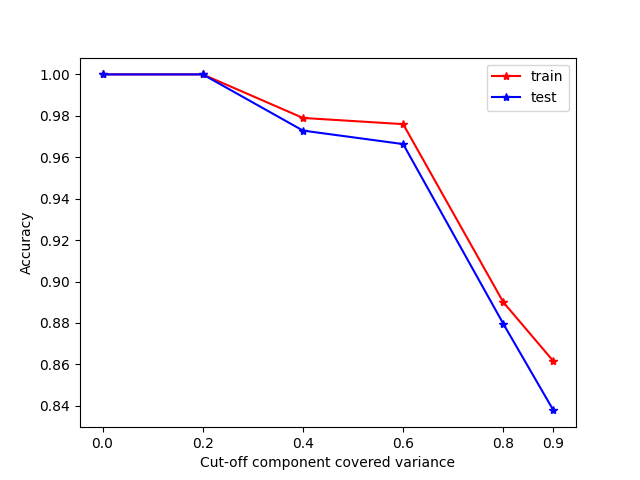}
\subcaption{Fashion MNIST, weaker XGB model (50 trees, depth 3)}
\end{minipage}
\begin{minipage}[t]{0.49\linewidth}
\centering
\includegraphics[width=8cm]{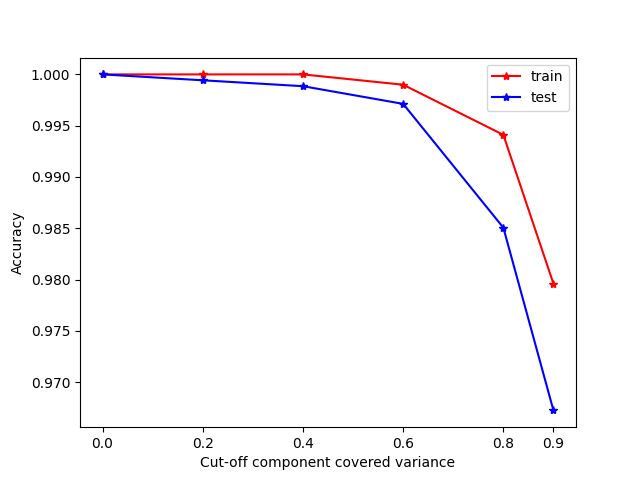}
\subcaption{Fashion MNIST, weaker MLP model (hidden layer size = 10)}       
\end{minipage}
\caption{Accuracy curves demonstrating the underfitting checking process: each graph shows the training accuracy and testing accuracy along with the percentage of data variance being removed
Upper two graphs a) and b) correspond to the underfitting scenarios we aim at, since training accuracy drops together with the testing accuracy.
The middle two graphs c) and d) show an overfitting situation, where testing accuracy drops quickly while training accuracy remains close to 100\%, and the gap between the two becomes larger and larger. This case should be avoided by choosing a weaker base model in the input feature space $X$. Graphs e) and f) then show that a weaker base model on the same dataset as c) and d) leads to underfitting when compared to the overfitting base model.}
\label{fig:accu_curve}
\end{figure*}

From here we use two procedures to generate diagnosis labels for miss-classified data: 

\begin{enumerate}
    \item Lower the capacity of the model. For example, for an XGBoost model decrease the number of trees and their depth. For an MLP reduce the number of hidden layers and neurons, or simply add noise to the weights. When training a lower capacity model on the cleaned-up data, some of the test sample will be miss-classified, and we can assign them the label ``Weak Model''.

\item In the input space $X$, drop some of the components of the features, which is equivalent to projecting the data to a lower dimensional vector space. This entails information loss, and classes that were well separated in the input space $X$ may overlap (mix up) in the projection space $X'$. Then train a strong classifier in the $X'$ space, without overfitting it. This classifier, because some classes are mixed-up in the reduced feature space, cannot achieve perfect accuracy, so there are some miss-classified data points on the test set. For this reason, these miss-classified data points can be labeled as ``Data Mixed-up''. Because strong classifiers are still very good even after throwing away a lot of features, at least on the datasets we are using, we actually perform a principal component analysis in the input space $X$ and then eliminate the most important components (to quickly remove components of high variance).
In order to achieve the desired effect, it is important to train the best model on the projected data \emph{without overfitting}, since because the trained classifier is strong, overfitting is indeed a risk. Fig.~\ref{fig:accu_curve} illustrates a situation when both MLP and XGBoost classifiers overfit depending on the dataset. If a classifier overfits when dropping data components, that indicates that a weaker version of it should be used to avoid this phenomenon.
\end{enumerate}

\subsection{Meta-classifier: training and prediction}
\label{sec:resmeta}

Each dataset is processed by the proposed method to extract meta-feature profile vectors for each element, together with the associated diagnosis labels, the result being a three class classification vector set. After the usual train/test dataset split we use a decision tree as our meta-classifier, because this type of model can be used to generate interpretable decision rules easily. 
A decision tree can also easily deal with large training sets, which is an important factor in our case, because we employ several large datasets to build the ground-truth for the meta-classifier. Meta-feature extraction time grows linearly with the input dataset size, but this process is only performed once.

The meta-classifier is evaluated using the accuracy, precision and recall scores. Our meta-classifier is capable of suggesting a diagnosis label for a new data point, miss-classified  by a different classifier on a different dataset and, if desired, decision rules extracted from the meta-classifier. These should help understanding if the data point has been wrongly classified because the classifier was too weak or the classification classes are mixed up in the feature space, and the decision rule gives insight into what happens locally in the feature space around the miss-classified point.

\subsection{Evaluation results}
\label{sec:reseval}

To test our proposal we use eight datasets set up for binary classification: MNIST 3-8~\cite{mnist} (classes 3 and 8, number of features reduced to 87 by PCA), Fashion MNIST 0-6~\cite{fashionmnist} (classes 0 and 6, number of features  reduced to 100 by PCA), Weather Australia\footnote{ \href{http://www.bom.gov.au/climate/dwo/}{http://www.bom.gov.au/climate/dwo}}, Spotify\footnote{ \href{https://github.com/rfordatascience/tidytuesday/tree/master/data/2020/2020-01-21}{https://github.com/rfordatascience/tidytuesday/tree/master/data/2020}}, Human Resource Analytics\footnote{\href{https://www.kaggle.com/datasets/rohandx1996/human-resource-analytics}{https://www.kaggle.com/datasets/rohandx1996/human-resource-analytics}}, 
Water Potability\footnote{\href{https://www.kaggle.com/datasets/adityakadiwal/water-potability}{https://www.kaggle.com/datasets/adityakadiwal/water-potability}}, Indian Diabetes\footnote{\href{https://www.kaggle.com/datasets/uciml/pima-indians-diabetes-database}{https://www.kaggle.com/datasets/uciml/pima-indians-diabetes-database}}, Banknote authentication\footnote{\href{https://archive.ics.uci.edu/ml/datasets/banknote+authentication}{https://archive.ics.uci.edu/ml/datasets/banknote+authentication}}.

On each dataset we train two types of classifiers: an XGBoost and a Multi Layer Perceptron (MLP). Each of them provides three models: the base model (obtained on the cleaned-up dataset (also called ``easy dataset'' in the following), as explained in Sec.~\ref{sec:resgt}), the weak model (obtained by weakening the base model either by reducing the number of trees and their depth, for XGBoost, or by reducing the size of the hidden layer, for the MLP), and the truncated feature space model (obtained after cutting off a number of components from the feature space).

In the Table~\ref{tab:data} we present the details of the involved datasets and configurations during the diagnosis label generation process: the dataset size including number of label 0 and label 1 points, the number of features, the number of extracted meta-vectors (corresponding to miss-classified samples), the parameters of baseline models on the easy (cleaned up) dataset, parameters of the weak model (for ``Weak Model'' label generation) and the number of cut-off data feature components (for ``Data Mixed-up'' label generation). For the ``MLP cut'' we also give the variance of the cut-off components (percentage in brackets). Notice that the first five datasets each provide 2000 meta-vectors, while the last three datasets, which are much smaller, give roughly ten times less meta-vectors.

Table~\ref{tab:mclf} presents results for four main configurations on extracted meta-vectors to train the meta-classifier decision tree and test its performance:
\begin{itemize}
    \item Row 1: train on 4 datasets randomly sampled from  \#1 to \#5 and test on datasets \#6 to \#8.
    \item Row 2: train on 4 randomly selected datasets and test on the rest.
    \item Row 3: train on XGBoost generated meta-vectors, test on MLP generated meta-vectors and vice-versa.
   \item Row 4: train/test on the union of all datasets: split 75\% train / 25\% test.
\end{itemize}

These configurations aim to test the ability of the meta-classifier to generalize across different datasets (train on a collection of datasets and predict on a different dataset: rows 1 and 2) and also to generalize across different classifiers (train on a classifier and predict on another classifier: row 3). The last configuration (row 4) represents the case where training and testing are done on the union of all the datasets and, as expected, provides slightly better results compared to the other configurations because the training dataset is much larger.

The decision tree structure, including tree depth, number of leaves, training set and testing set size, most important meta-features from the tree, and particularly the most impacting decision rules extracted from the tree, are also shown in Table~\ref{tab:mclf}. Performance of the meta-classifiers is provided by predicting precision and recall scores on the test set (each contains three elements corresponding to each of the three diagnosis classes), where the precision/recall values are collected statistically from several rounds 
of training, giving the mean and standard derivation. 

\begin{table*}[t]
\centering
\begin{tabular}{|l|c|c|c|c|c|c|c|}
    \hline
    Dataset & \# of data points & \# of  & \# of  & XGB base & XGB weak & MLP base & MLP weak \\
     & \#label0, \#label1 & features & meta-vectors & easyset size & XGB cut & easyset size & MLP cut \\
    \hline
    1) MNIST 3-8 & 11982 & 87 & xgb 2000 & 50 trees, depth 3 & 40 trees, depth 1 & 100 iter, (200,)hl & 5 iter (10,)hl\\
     & 6131, 5851 &  & mlp 2000 & 10343 & 40 comp (90\%) & 11153 & 70 comp (98\%)\\
    \hline
    2) Fashion MNIST 0-6 & 12000 & 100 & xgb 2000 & 50 trees, depth 3 & 10 trees, depth 1 & 100 iter (500,)hl & 7 iter (4,)hl\\
     & 6000, 6000 &  & mlp 2000 & 9214 & 30 comp (80\%) & 8699 & 12 comp (50\%) \\
    \hline
    3) Weather AUS & 20000 & 17 & xgb 2000 & 100 trees, depth 3 & 10 trees, depth 2 & 500 iter (30,)hl & 7 iter (4,)hl \\
     & 15669, 4331 &  & mlp 2000 & 16114 & 3 comp (60\%) & 16062 & 6 comp (80\%) \\
    \hline
    4) Spotify Pop & 32833 & 12 & xgb 2000 & 120 trees, depth 4 & 50 trees, depth 2 & 500 iter (100,)hl & 7 iter (10,)hl \\
     & 18871, 13962 &  & mlp 2000 & 18551 & 6 comp (60\%) & 14762 & 4 comp (50\%) \\
    \hline
    5) Human Resource & 14999 & 5 & xgb 2000 & 100 trees, depth 4 & 10 trees, depth 1 & 1000 iter (100,)hl & 20 iter (5,)hl \\
     & 11428, 3571 &  & mlp 2000 & 14512 & 3 comp (60\%) & 14441 & 3 comp (60\%) \\
    \hline
    \hline
    6) Water Potability & 3276 & 9 & xgb 200 & 20 trees, depth 4 & 10 trees, depth 1 & 1000 iter (50,)hl & 50 iter (10,)hl \\
     & 1998, 1278 &  & mlp 500 & 1994 & 5 comp (60\%) & 1769 & 5 comp (60\%) \\
    \hline
    7) Diabetes & 767 & 8 & xgb 194 & 25 trees, depth 3 & 10 tree3, depth 1 & 200 iter (50,)hl & 100 iter (5,)hl \\
     & 500, 267 &  & mlp 200 & 534 & 3 comp (60\%) & 568 & 3 comp (60\%) \\
    \hline
    8) Banknote & 1371 & 4 & xgb 500 & 50 trees, depth 3 & 10 trees, depth 2 & 1000 iter (40,)hl & 60 iter (5,)hl \\
     & 761, 610 &  & mlp 500 & 1354 & 2 comp (60\%) & 1366 & 2 comp (60\%) \\
    \hline
\end{tabular}
\caption{Information about the used datasets and the parameters of base models, weak models and cut-off models. All MLP models have one hidden layer (its size is given by the HL parameter). For the XGBoost model, we give the number of trees and their depth. See Sec.~\ref{sec:reseval} for a detailed description of each column.}
\label{tab:data}
\end{table*}

\begin{table*}[t]
    \centering
    \resizebox{\linewidth}{!}{%
    \begin{tabular}{|c|c|c|c|c|c|c|c|}
    \hline
        Data  & \# of  & Dtree & Dtree & Important &\textbf{Extracted} & \# of & Prediction  \\
        Configuration & train vectors & depth & leafs & meta features &\textbf{Decision Rules} & test vectors &  \textbf{Precision / Recall} \\
        \hline
        train  & 11075 & 16 & 188 & rTN 0.31, &WM:  (rFN$>$0.17)\&(rate dist gt$<=$0.435) & 1659 & Prec: [0.83±0.0465  \\
        4 datasets &&&& rate dist gt 0.27, & & & 0.978±0.0144  \\
        from 1)-5); &&&& rate dist pred 0.14 &C:  (rTN$<=$0.03)\&(rate dist pred$<=$0.493) & & 0.828±0.0431] \\
        test 6)-8) &&&& proximity 0.07 & &&Rec: [0.854±0.0688 \\
         &&&& knn pred conf 0.05 &MD:  (rFN$>$0.03)\&(knn pred conf$>$0.097) &&0.932±0.011 \\
         &&&&&\&(rate dist gt$>$0.501)\&(rate dist pred$<=$0.381) && 0.838±0.0648] \\
        \hline
        train random  & 12450 & 16 & 153 & rTN 0.39, & WM:  (rTN$>$0.63)\&(rate dist gt$<=$0.533) & 16560 & Prec: [0.766 ±0.0736 \\
        4 datasets; &&&& rate dist gt 0.24, &\&(MST frac gt$>$0.02)\&(rate dist pred$<=$0.509) & & 0.964 ±0.02 \\
        test rest  &&&& rate dist pred 0.10, &C:  (rTN$<=$0.01)\&(rate dist pred$<=$0.501)  && 0.774 ±0.0886 ] \\
        4 datasets &&&& knn pred conf 0.06, & & & Rec: [0.804 ±0.0952 \\
         &&&& MST frac gt 0.05 &MD:  (rTN$>$0.01)\&(rate dist gt$>$0.56) && 0.904 ±0.0548 \\
         &&&&&\&(rate dist pred$<=$0.39)\&(knn pred conf$>$0.09)  && 0.778 ±0.0835 ] \\
        \hline
        train all & 11050 & 18 & 163 & rFN 0.42, &WM:  (rFN$>$0.13)\&(rate dist gt$<=$0.437)& 12200 & Prec:  [0.7575 ±0.0538  \\
        XGB vec; &&&& rate dist gt 0.22, &\&(MST frac gt$>$0.112)\&(local set cardinality pred$>$0.078)&& 0.9425 ±0.0471 \\
        test all &&&& rate dist pred 0.14, &C:  (rFN$<=$0.03)\&(rate dist pred$<=$0.498)&& 0.7675 ±0.0316 ] \\
        MLP vec. &&&& proximity 0.06, &\&(rTN$<=$0.68)&& Rec: [0.8075 ±0.0983 \\
        and vice &&&& rTN 0.04 &MD:  (rFN$>$0.03)\&(rate dist gt$>$0.467)&& 0.9 ±0.0763 \\
        versa &&&& &\&(rate dist pred$<=$0.297)\&(proximity$<=$0.997)&& 0.7475 ±0.0792 ] \\
        \hline
        train all  & 11300 & 15 & 181 & rFN 0.31, &WM:  (rFN$>$0.15)\&(rate dist gt$<=$0.413) & 3885 & Prec: [0.9±0.00931 \\
        random split &&&& rate dist gt 0.26,&\&(MST frac gt$>$0.112) & & 0.972±0.011 \\
        75\%; &&&& rate dist pred 0.13 &C:  (rFN$<=$0.03)\&(rate dist pred$<=$0.49) & & 0.882 ±0.0144] \\
        test rest  &&&& proximity 0.06 &\&(rTN$<=$0.19) && \\
        25\% &&&& knn pred conf 0.05 &MD:  (rFN$>$0.03)\&(rate dist gt$>$0.501) &&Rec: [0.908 ±0.011 \\
        &&&&&\&(rate dist pred$<=$0.37)\&(knn pred conf$>$0.096) &&0.92 ±0.0246 \\
        &&&&&\&(MST frac gt$<=$0.459) && 0.898 ±0.0172] \\
        \hline
    \end{tabular}%
    }
    \caption{Learned meta classifier (decision trees), their properties and predicting performance scores. }
    \label{tab:mclf}
\end{table*}

\begin{figure}[htbp]
\centering
\includegraphics[width=\linewidth]{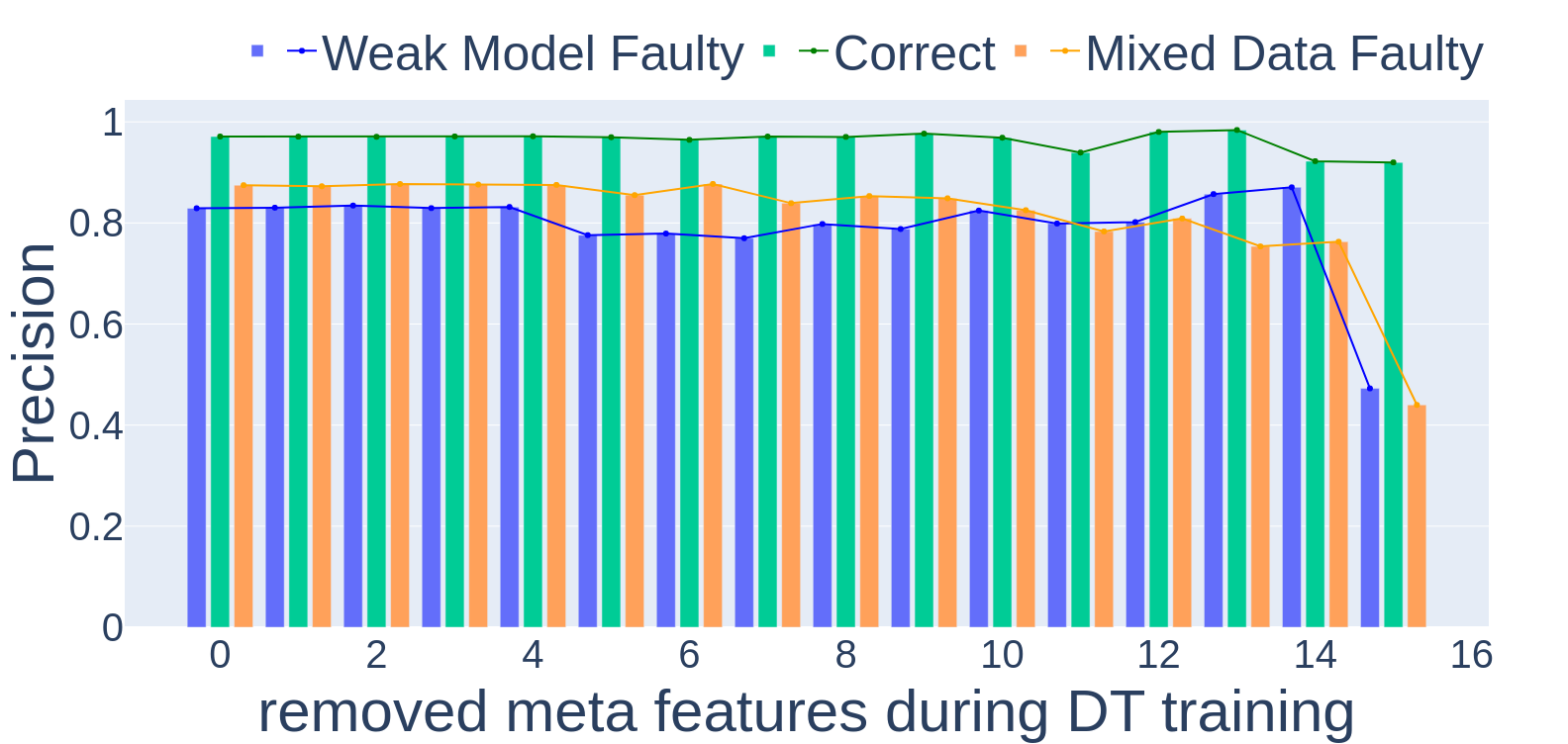}
\vspace{5pt}
\includegraphics[width=\linewidth]{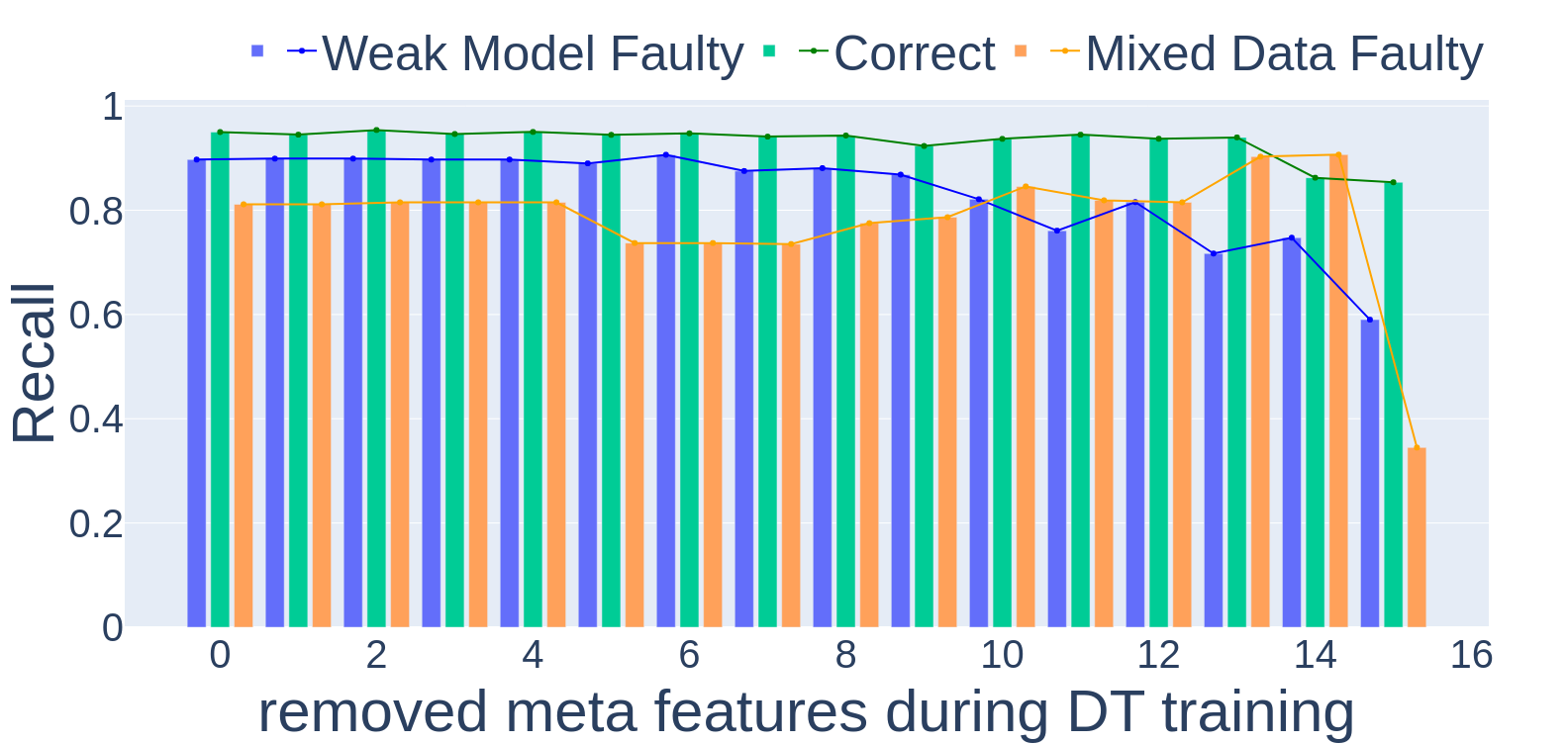}
\caption{Linked Bar plots of meta feature ablation experiment, showing that prediction accuracy score decays with important meta-feature removed (from least important ones to more important ones).}
\label{fig: prec}
\end{figure}

Looking at Table~\ref{tab:mclf}, for all experimental configurations, after proper label re-balancing on training vectors, the obtained decision trees have similar structure: the number of features with a high score according to the decision tree feature importance score is quite low (usually less than 5 or 6), although the specific features may vary depending on the configuration. Some of the meta-features are always present and provide significant discriminative power on three-diagnosis-labels classification task (e.g.\ \textit{rTN/rFN} -- rate of True Negative/False Negative data, \textit{rate dist gt/pred} -- rate of average distance to ally neighbors against opponent neighbors in terms of ground truth labels/model predicted labels, \textit{KNN pred conf} -- KNN average predicted probability confidence measure; see Sec.~\ref{sec:metafeatures} for a detailed description of these features). They are thus good candidates to form a small meta-feature set in terms of which to present the decision rules.

Most meta-classifiers have \emph{precision and recall values around 80\%--90\%} on all three diagnosis labels, showing that our method provides a high degree of invariance over the dataset and classifier type. This means that we can learn the meta-classifier on a collection of datasets and predict diagnosis labels for different datasets and classifiers.

In a few cases a wee see a higher uncertainty level (accuracy drop under 80\%). To explain this performance drop, one reason might be the lack of enough training meta-vectors (see, for example, Table \ref{tab:mclf} row 2) causing decision tree to overfit on the training data, or the different geometrical nature of decision boundaries from different types of model (see Table \ref{tab:mclf} row 3). This indicates that in order to improve the meta-classifier with respect to the pipeline performance, more datasets of diverse types and  complexity are needed.

Based on previous findings we performed an \textit{ablation test} (see Fig.~\ref{fig: prec}). We remove meta-features in the order of their increasing importance score (starting with the smallest but non-zero value): as more and more meta-features are removed (actually being made zero thus muted during training), we see the evolution of precision and recall value, on the three diagnosis classes. The curves are stable up until dropping the last 5 or 6 meta-features, indicating that indeed, these most important meta-features contain the majority of the discriminating power in terms of the three diagnosis classes, while the rest of the meta-features hold for the ability to gain higher predictive accuracy on more granular patterns in meta-feature space.

Concerning the computing resources, to perform the evaluations presented here, we used a standard personal computer with a 2.5GHz Intel Core i5-12600HX CPU (12 cores 16 threads) and 16GB of RAM. The most expensive part is the generation of the meta-features, which for the largest dataset used here (Spotify Pop) took around 20 min to obtain a total of 4000 meta-vectors. This time grows linearly with the size and the number of features of the dataset but can be reduced by using a more sophisticated KNN retrieval algorithm like the one in \cite{johnson2019billion}. Training the meta-classifier took a few minutes using a standard Scikit-learn\footnote{ \href{https://scikit-learn.org}{https://scikit-learn.org}} decision tree implementation.

\section{Conclusion}
\label{sec:conclusion}

In this work we propose a framework to help the user gain information about why a classifier failed to give a correct result for a given data point (given individual query). We introduce several quantities describing what happens around the faulty point in the feature space and, with the aid of these, we extract profile vectors as intermediate representations capable of unifying faulty data from different datasets and different classifiers. These profile vectors are aggregated by a decision tree meta-classifier to match three diagnosis cases. The final interpretation rules provided by the decision tree could be useful to help the user distinguish and understand the reason of faulty data. 

Users may employ the proposed set of interpretability tools to debug an ML model: decide whether they should add more data or more features, prune the model, or fine-tune other hyper-parameters inside the model, in order to achieve higher prediction accuracy and better generalisation, or else just correct certain parts of the dataset to remove ambiguity.

One obvious direction in which to develop the presented method is to extend the framework to include the overfitting case. The difficulty comes from the fact that in a region where several classes overlap, a weak classifier will underfit, while a strong one will overfit, both leading to poor generalisation. We may need in this case a finer graduation of diagnosis scenarios, which might not be feasible in all situations.

An important development direction of our proposal would be to extend it to work in the case of \textit{multi-class classifiers}. Indeed, in the present work we tested our framework in the case of \textit{binary classifiers}, but there is nothing that limits it to this: the definition of a miss-classified sample remains the same (a sample that that is attributed by the classifier a wrong label) and the construction of the ground-truth for the meta-classifier can be done by the procedure described in Sec.~\ref{sec:resgt} in a one-vs-rest manner. The meta-features list should probably be updated to include a measure of the degree of local non-separability of different pairs of classes in the neighbourhood of the test point. We also expect a linear increase in the scale of certain meta-features calculation, such as the confusion matrix, since there should be an element for each class.

Another direction worthy of investigation is to develop the meta-features automatically, for example by using a neural network to project the input space into an intermediate representation that maximizes separability between the clusters associated with each diagnosis label. This might indeed lead to better meta-classifiers but would likely require a much larger training set and, in addition, would lose the interpretability of the meta-features.

\bibliographystyle{unsrt}
\bibliography{mybib}

\end{document}